\documentclass[conference]{article}
\usepackage{spconf,amsmath,graphicx,subfigure,amssymb,flushend}
\usepackage{algorithmic, algorithm,color}
\usepackage{fancyheadings}
\title{Barcode Annotations for Medical Image Retrieval:\\ A Preliminary Investigation}

\name{H.R.Tizhoosh\thanks{The author would like to thank NSERC (Natural Sciences and Engineering Research Council of Canada) for funding this project (Discovery Grant).}}
\address{Centre for Bioengineering and Biotechnology, University of Waterloo \\ Waterloo, ON, Canada, tizhoosh@uwaterloo.ca}

\begin{document}
\maketitle

\begin{abstract}
This paper proposes to generate and to use barcodes to annotate medical images and/or their regions of interest such as organs, tumors and tissue types. A multitude of efficient feature-based image retrieval methods already exist that can assign a query image to a certain image class. Visual annotations may help to increase the retrieval accuracy if combined with existing feature-based classification paradigms. Whereas with annotations we usually mean textual descriptions, in this paper \textbf{barcode annotations} are proposed. In particular, Radon barcodes (RBC) are introduced. As well, local binary patterns (LBP) and local Radon binary patterns (LRBP) are implemented as barcodes. The IRMA x-ray dataset with 12,677 training images and 1,733 test images is used to verify how barcodes could facilitate image retrieval. 
\end{abstract}


\section{Idea and Motivation}
\label{sec:intro}
The idea proposed in this paper is to generate short barcodes, embed them in the medical images (i.e. in DICOM files) and use them, along with feature-based methods, for fast and accurate image search. Retrieving (similar) medical images when a query image is given could assist clinicians for more accurate diagnosis by comparing with similar (retrieved) cases. As well, image retrieval can contribute to research-oriented tasks in biomedical imaging in general (e.g. in histopathology). 
 
 Why not features? This proposal does not seek to replace feature-based classification approach to content-based image retrieval (CBIR). Paradigms such as ``bag of words'' and ``bag of features'' along with powerful classifiers such as SVM and KNN have undoubtedly moved the research forward in image search. Barcode annotations should solely assist in this process, particularly for medical images, as additional source of information. Even though the retrieval capabilities of barcodes will be examined in this paper, they are not meant to be the main retrieval mechanism but auxiliary to existing ones.   
 
 Why barcodes? As supplementary information, barcodes could enhance the results of existing ``bag-of-features'' methods, which are generally designed to capture the global appearance of the scene without much attention to the local details of scene objects (e.g. shape of a tumor in an MR scan). Specially local, ROI-based barcodes may be more expressive in capturing spatial information. Whether 1D or 2D barcodes are used may deliver the same results even though 1D barcodes are expected to be shorter and hence faster in execution.  

\section{Literature review}
Since Radon barcodes will be introduced in this paper, and because local binary patterns are implemented for sake of comparison, in following the relevant literature will be briefly reviewed. Due to space limitations we cannot review the vast literature on CBIR as adequately as we generally do.

\textbf{Literature on Radon transform --} Depicting a three dimensional object is the main motivation for Radon transform. There are many applications of Radon transform reported in literature. Zhao et al. \cite{zhao2013} use feature detection in Radon transformed images to calculate the ocean wavelength and wave direction in SAR images. Hoang and Tabbone \cite{hoang2012} employed Radon transform in conjunction with Fourier and Mellin transform for extraction of invariant features. Nacereddine et al. \cite{nacereddine2010} also used Radon transform to propose a new descriptor called Phi-signature for retrieval of simple binary shapes. Jadhav and Holambe \cite{jadhav2009} use Radon transform along with discrete wavelet transform to extract  features for face recognition. Chen and Chen \cite{chen2008} introduced Radon composite features (RCFs) that transform binary shapes into 1D representations to calculate features from. Tabbone et al. \cite{tabbone2008} propose a histogram of the Radon transform (HRT), which is invariant to geometrical transformations. They use HRT for shape retrieval. Dara et al. \cite{daras2006} generalized Radon transform to radial and spherical integration to search for 3D models of diverse shapes. Trace transform is also a generalization of Radon transform \cite{kadyrov2001, kadyrov2004} for invariant features via tracing lines applied on shapes with complex texture on a uniform background for change detection. Although binary images/thumbnails have been used to facilitate image search \cite{PatentConsensus,PatentQB,Daugman2004,Arvancheh2006}, it seems that no attempt has been made to binarize the Radon projections and use them directly for CBIR tasks as it will be proposed in this paper.  
 
\textbf{Literature on LBPs and CBIR --} Local binary patterns (LBPs) were introduced by Ojala et al. \cite{ojala2002}. Among others, LBPs have been used for face recognition \cite{ahonen2006,galoogahi2012}. The literature on CBIR and medical CBIR is vast. Ghosh et al. \cite{ghosh2011} review online systems for content-based medical image retrieval such as GoldMiner, BioText, FigureSearch, Yottalook, Yale Image Finder, IRMA and iMedline. Multiple surveys are available that review recent literature  \cite{shandilya2010,rajam2013,dharani2013}. 

\section{Radon barcode annotations}
\label{sec:RBC}
\begin{figure}[htbp]
\begin{center}
\vspace{0.05in}
\includegraphics[width=0.90\columnwidth]{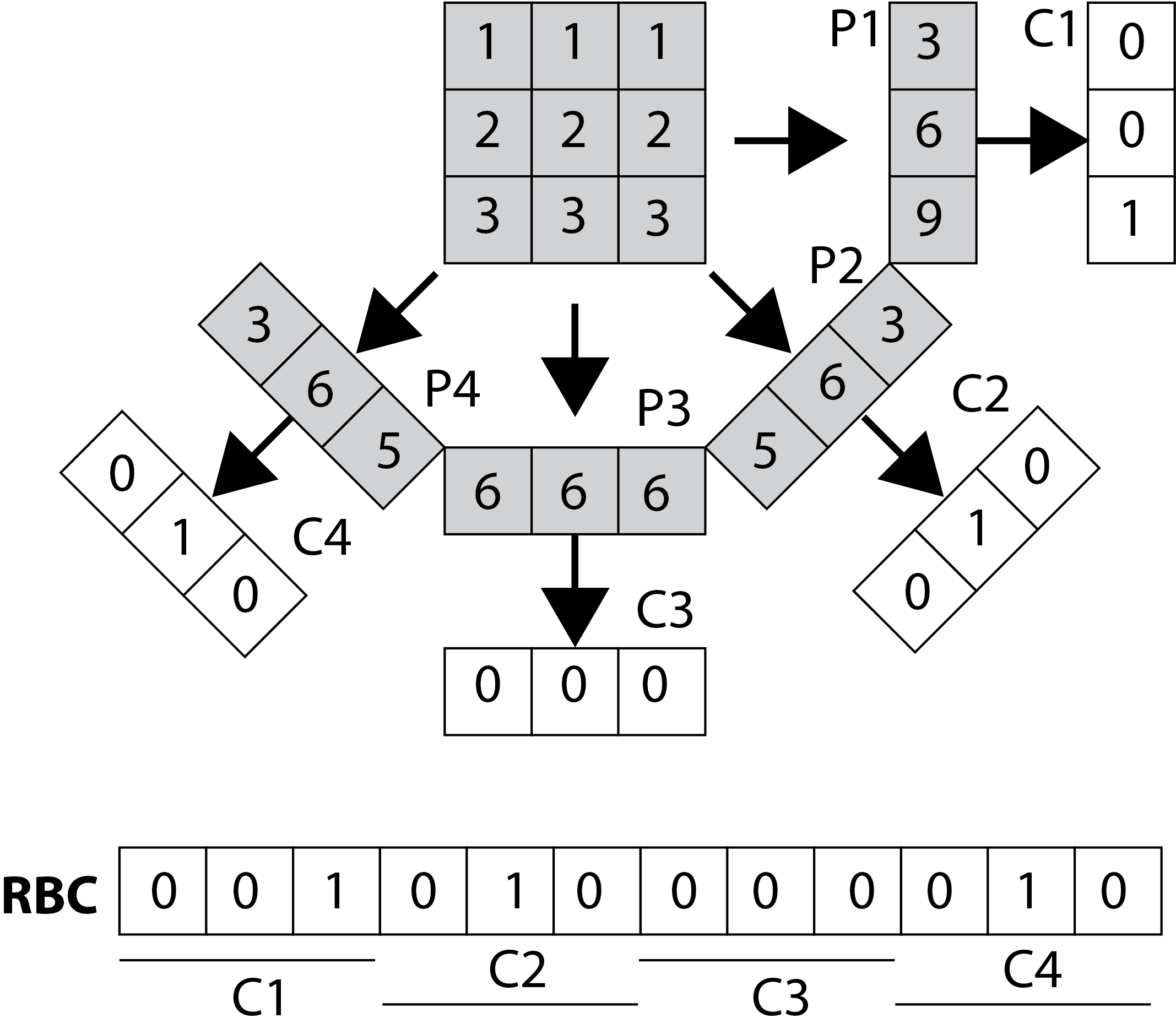}
\caption{Radon Barcode (RBC) -- The image is Radon-transformed. All projections (P1,P2,P3,P4) are thresholded to generate code fragments C1,C2,C3,C4. The concatenation of all code fragments delivers the barcode \textbf{RBC}. }
\label{fig:RBC}
\end{center}
\end{figure}

Examining a function $f(x,y)$, one can project $f(x,y)$ along a number of projection angles. The projection is basically the sum (integral) of $f(x,y)$ values along lines constituted by each angle $\theta$. The projection creates a new image $R(\rho,\theta)$ with $\rho = x \cos \theta + y \sin \theta$. Hence, using the Dirac delta function $\delta(\cdot)$ the Radon transform can be written as 
\begin{equation}
R(\rho,\theta) = \int\limits_{-\infty}^{+\infty} \int\limits_{-\infty}^{+\infty} f(x,y) \delta(\rho-x\cos \theta-y\sin\theta) dx dy.
\end{equation}
If we threshold all projections (lines) for individual angles based on a ``local'' threshold for that angle, then we can assemble a barcode of all thresholded projections as depicted in Figure \ref{fig:RBC}. A simple way for thresholding the projections is to calculate a typical value via median operator applied on all non-zero values of each projection. Algorithm \ref{alg:Radon} describes how \textbf{Radon barcodes (RBC)} are generated \footnote{Matlab code available online: http://tizhoosh.uwaterloo.ca/}.  In order to receive same-length barcodes \emph{Normalize$(I)$} resizes all images into $R_N\times C_N$ images (i.e. $R_N= C_N=2^n,n\in \mathbb{N}^+$).

\begin{algorithm}[htbp]
\caption{Radon Barcode (RBC) Generation}
\begin{algorithmic}[1]
\label{alg:Radon}
\STATE Initialize Radon Barcode $\mathbf{r} \leftarrow \emptyset$ 
\STATE Initialize angle $\theta \leftarrow 0$ and $R_N=C_N\leftarrow 32$
\STATE Normalize the input image $\bar{I} = \textrm{Normalize}(I,R_N,C_N)$ 
\STATE Set the number of projection angles, e.g. $n_p \leftarrow 8$
\WHILE{$\theta < 180$}
	\STATE Get all projections $\mathbf{p}$ for $\theta$
	\STATE Find typical value $T_\textrm{typical}\leftarrow\textrm{median}_i (\mathbf{p}_i)|_{\mathbf{p}_i \neq 0}$
	\STATE Binarize projections: $\mathbf{b} \leftarrow \mathbf{p} \geq T_\textrm{typical}$ 
	\STATE Append the new row $\mathbf{r} \leftarrow \textrm{append}(\mathbf{r},\mathbf{b} )$ 
	\STATE $\theta \leftarrow \theta + \frac{180}{n_p}$
\ENDWHILE
\STATE Return  $\mathbf{r}$
 \end{algorithmic}
 \end{algorithm}

\section{LBP Barcode Annotations}
\begin{figure}[htb]
\centering     
\subfigure[input image]{\label{fig:a}\includegraphics[width=30mm,height=30mm]{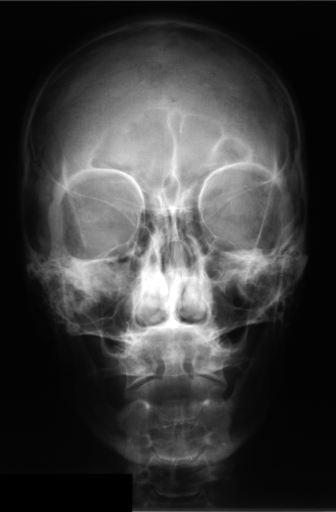}}\qquad
\subfigure[input image]{\label{fig:a}\includegraphics[width=30mm,height=30mm]{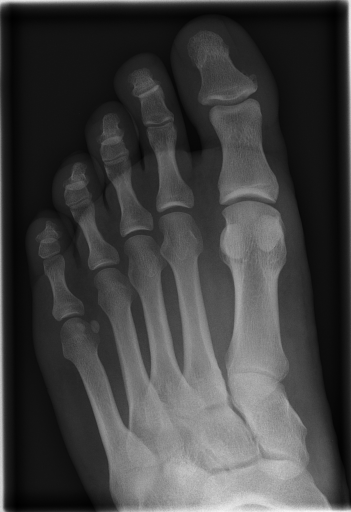}}\\
\subfigure[RBC$_8$]{\label{fig:a}\includegraphics[width=0.4\columnwidth,height=5mm]{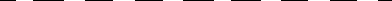}}\qquad
\subfigure[RBC$_8$]{\label{fig:a}\includegraphics[width=0.4\columnwidth,height=5mm]{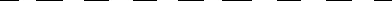}}\\
\subfigure[RBC$_{16}$]{\label{fig:a}\includegraphics[width=0.4\columnwidth,height=5mm]{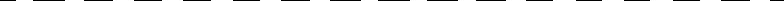}}\qquad
\subfigure[RBC$_{16}$]{\label{fig:a}\includegraphics[width=0.4\columnwidth,height=5mm]{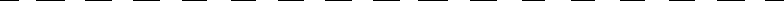}}\\
\subfigure[RBC$_{32}$]{\label{fig:a}\includegraphics[width=0.4\columnwidth,height=5mm]{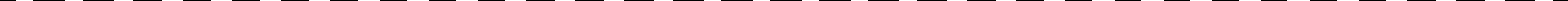}}\qquad
\subfigure[RBC$_{32}$]{\label{fig:a}\includegraphics[width=0.4\columnwidth,height=5mm]{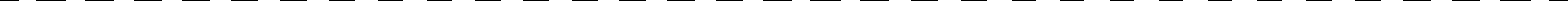}}\\
\subfigure[LBP]{\label{fig:a}\includegraphics[width=0.4\columnwidth,height=5mm]{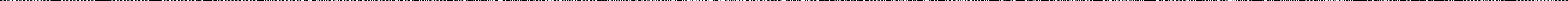}}\qquad
\subfigure[LBP]{\label{fig:a}\includegraphics[width=0.4\columnwidth,height=5mm]{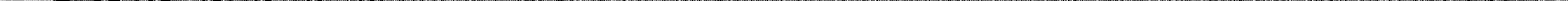}}\\
\subfigure[LRBP$_4$]{\label{fig:a}\includegraphics[width=0.4\columnwidth,height=5mm]{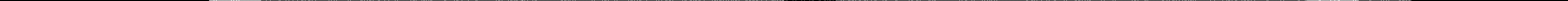}}\qquad
\subfigure[LRBP$_4$]{\label{fig:a}\includegraphics[width=0.4\columnwidth,height=5mm]{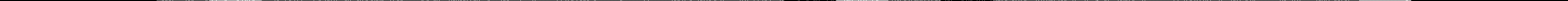}}\\
\subfigure[LRBP$_{32}$]{\label{fig:a}\includegraphics[width=0.4\columnwidth,height=5mm]{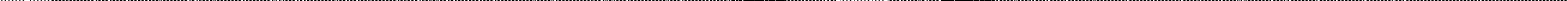}}\qquad
\subfigure[LRBP$_{32}$]{\label{fig:a}\includegraphics[width=0.4\columnwidth,height=5mm]{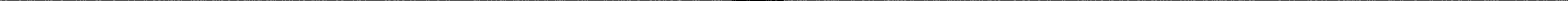}}\\
\caption{Sample barcodes for RBC, LBP and LRBP}
\label{fig:Barcodes}
\end{figure}

Local binary patterns (LBPs) have been extensively used in image classification. The most common usage of LBPs is to calculate their histogram and use them as features. Here, for sake of comparison, LBPs are extracted and recorded as barcodes through concatenation of binary vectors around each pixel. As well, the Galoogahi and Sim approach \cite{galoogahi2012} to apply LBP on Radon transformed images, namely local Radon binary patterns (LRBPs), is also implemented to compare with the proposed RBC. Figure \ref{fig:Barcodes} shows barcode annotations for two medical images from IRMA dataset (see section \ref{subsec:IRMA}) for different $n_p$ values (see lines 4 and 10 in Algorithm \ref{alg:Radon}).

\section{Experiments}
\subsection{Image Test Data}
\label{subsec:IRMA}
\begin{figure*}[htb]
\centering     
\subfigure[1121-127-700-500]{\label{fig:a}\includegraphics[width=28mm,height=28mm]{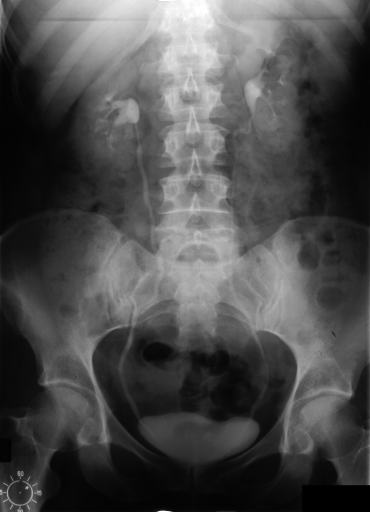}}
\subfigure[1121-110-414-700]{\label{fig:b}\includegraphics[width=28mm,height=28mm]{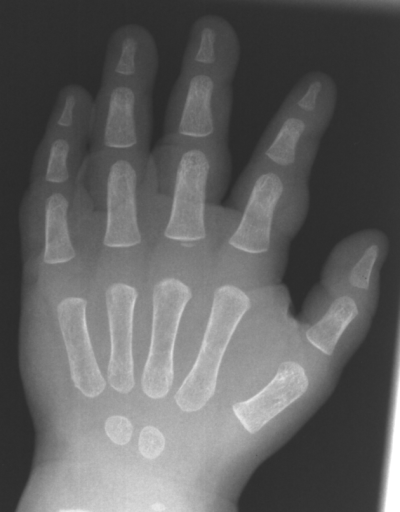}}
\subfigure[1121-120-942-700]{\label{fig:b}\includegraphics[width=28mm,height=28mm]{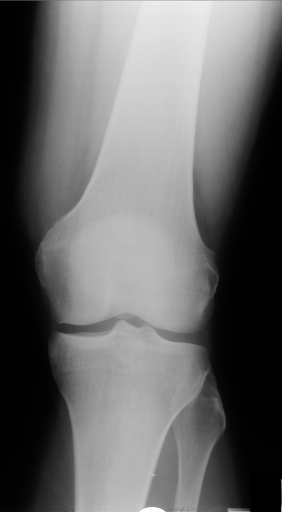}}
\subfigure[1123-127-500-000]{\label{fig:b}\includegraphics[width=28mm,height=28mm]{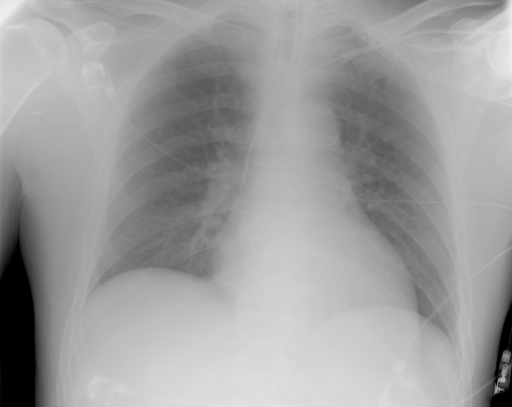}}
\subfigure[1121-120-200-700]{\label{fig:b}\includegraphics[width=28mm,height=28mm]{4068.png}}\\
\subfigure[1121-200-412-700]{\label{fig:b}\includegraphics[width=28mm,height=28mm]{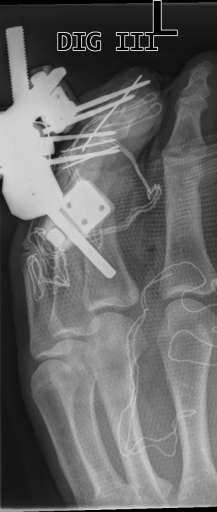}}
\subfigure[1121-120-918-700]{\label{fig:b}\includegraphics[width=28mm,height=28mm]{9856.png}}
\subfigure[1121-240-442-700]{\label{fig:b}\includegraphics[width=28mm,height=28mm]{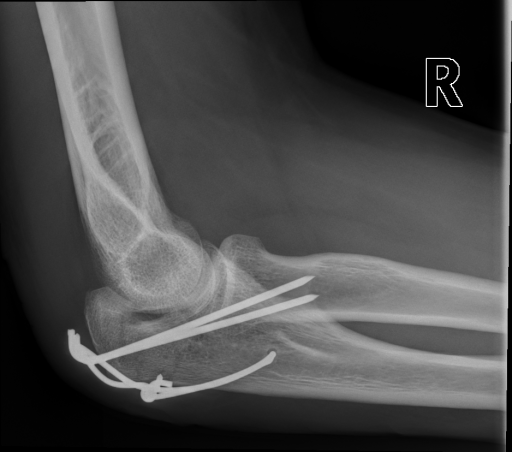}}
\subfigure[1121-220-310-700]{\label{fig:b}\includegraphics[width=28mm,height=28mm]{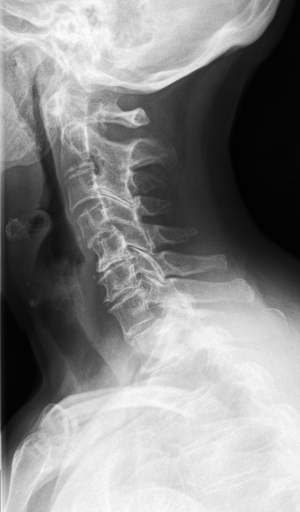}}
\subfigure[112d-121-500-000]{\label{fig:b}\includegraphics[width=28mm,height=28mm]{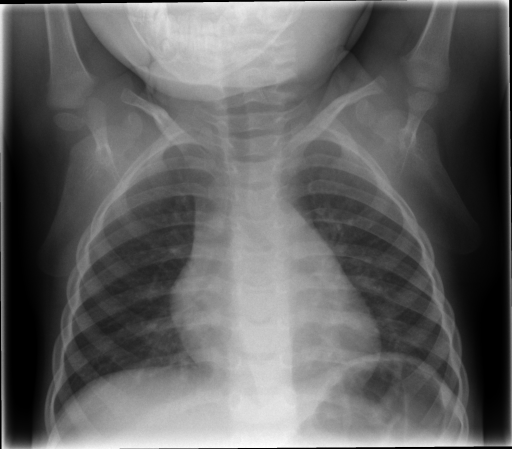}}
\caption{Sample images from IRMA Dataset with their IRMA codes TTTT-DDD-AAA-BBB.}
\label{fig:IRMASamples}
\end{figure*}

The Image Retrieval in Medical Applications (IRMA) database\footnote{http://irma-project.org/} is a collection of more than 14,000 x-ray images (radiographs) randomly collected from daily routine
work at the Department of Diagnostic Radiology of the RWTH Aachen University\footnote{http://www.rad.rwth-aachen.de/} \cite{Lehmann2003,Mueller2010}. All images are classified into 193 categories (classes) and annotated with the IRMA code which relies on class-subclass relations to avoid ambiguities in textual classification \cite{Mueller2010,Lehmann2006}. The IRMA code consists of four mono-hierarchical axes with three to four digits each: the technical code T (imaging modality), the directional code D (body orientations), the anatomical code A (the body region), and the biological code B (the biological system examined). The complete IRMA code subsequently exhibits a string of 13 characters, each in $\{0,\dots,9;a,\dots,z\}$:\\
 \begin{equation}
 \textrm{TTTT-DDD-AAA-BBB}. 
 \end{equation}
 More information on the IRMA database and code can be found in \cite{Lehmann2003,Lehmann2006,Mueller2010}. IRMA dataset offers 12,677 images for training and 1,733 images for testing. Figure \ref{fig:IRMASamples} shows some sample images from the dataset long with their IRMA code in the format  TTTT-DDD-AAA-BBB.

\subsection{Accuracy Measurements}
Similar to \cite{Mueller2010,Mueller2008}, the following equation is used to calculate the total error of image retrieval over 1,733 images each characterized through 4 IRMA codes with $n_d$ digits:
\begin{equation}
\label{eq:Rerror}
E_{\textrm{total}}(l^{\textrm{query}})=\sum_{i=1}^{1733} \sum_{k=1}^{4} \sum_{j=1}^{n_{d}} \frac{1}{B_j^{ik}} \frac{1}{j} \delta(l^{k,\textrm{query}}_j,l_j)
\end{equation}
where $n_d\in \{3,4\}$, and $B_j^{ik}=10$ assuming that every digit can have 10 different values $0,1,2,\dots,9$. The function $\delta(l^{k,\textrm{query}}_j,l_j)$ delivers 0 if $l^{k,\textrm{query}}_j=l_j$ otherwise 1.  As well, the total number of wrong digits in the IRMA codes of retrieved images are recorded in $n_{\textrm{wrong}}^{\textrm{digits}}$. Using the code length $L_{\textrm{code}}$, one can establish a suitability measure $\eta$ that prefers small mismatch, low error and short codes simultaneously:
\begin{equation}
\eta^k = \frac{\max\limits_i (n_{\textrm{wrong}}^{i,\textrm{digits}}) \times \max\limits_i (E_{\textrm{total}}^i) \times \max\limits_i  (L_{\textrm{code}}^i)}{n_{\textrm{wrong}}^{k,\textrm{digits}} \times E_{\textrm{total}}^k \times L_{\textrm{code}}^k}
\end{equation}
Apparently, the larger $\eta$, the better the method, a desired quantification if the code length is  important in computation. 

\subsection{Results}
 Experiments were performed with all different barcodes. The proposed Radon barcode (RBC) with different number of projection angles ($n_p=4,8,16,32$) was trained and tested. Among 12,677 IRMA images, 12,631 could be used for training (some images were ignored due to incomplete codes). The training basically annotates each image with a barcode. For testing, 1,733 IRMA images were used.  For each of the test images (that come with their IRMA codes) complete search was performed to find the most similar image whereas the similarity of an put image $I^{query}_{i}$ annotated with the corresponding barcode $\mathbf{b}^{query}_{i}$ was calculate based on Hamming distance to any other image $I_{j}$ with its annotated barcode $\mathbf{b}_{j}$:
 \begin{equation}
\underset{{j=1,2,\dots,1733,j\neq i}}{\textrm{argmax}} \quad 1 - \frac{|\textrm{XOR}(\mathbf{b}^{query}_{i},\mathbf{b}_{j})|}{|\mathbf{b}^{query}_{i}|}
 \end{equation}
 For sake of comparison, barcodes generated via local binary patterns (LBP) \cite{ojala2002} and local Radon binary pattern (LRBP) \cite{galoogahi2012} were also used to train and test image retrieval. All input images were resized to $32\times 32$ for all barcode approaches. LRBP was tested for two projections $n_p=4$ and $n_p=32$. Table \ref{table:results} shows all results. 
  
 \begin{table}[htdp]
\caption{Trained with 12,631 images and tested with 1,733 images, the number of wrongly retrieved IRMA code digits $n_{\textrm{wrong}}^{\textrm{digits}}$, the total error $E_{\textrm{total}}$ and the code length $L_{\textrm{code}}$ are reported. Barcodes are ranked based on their suitability $\eta$.}
\begin{center}
\begin{tabular}{|c|c|c|c||c|c|}
Annotation 	&	$n_{\textrm{wrong}}^{\textrm{digits}}$ 	&$E_{\textrm{total}}$  & $L_{\textrm{code}}$  & $\eta$ & Rank\\ \hline\hline
RBC$_4$	& 	33.1\%		& 476.62 & 512 &  15.526 & 1\\ 
RBC$_8$	& 	33.2\%		& 478.54 & 1024 & 7.708 & 2\\ 
RBC$_{16}$	& 	32.7\%		& 470.57 & 2048 & 3.979 & 3\\ 
RBC$_{32}$	& 	33.1\%		& 475.92 & 4096 & 1.944 & 4\\ \hline
LBP		& 	32.3\%		& 463.81 & 7200 & 1.163 & 5\\ \hline
LRBP$_4$	& 	33.8\%		& 483.54 & 7200 & 1.066 & 6\\
LRBP$_{32}$	& 	34.7\%		& 501.96 & 7200 & 1.000 & 7\\ \hline
\end{tabular}
\end{center}
\label{table:results}
\end{table}%

Looking at the results of the Table \ref{table:results}, one can state:
\begin{itemize}
\item RBC with 4 Radon projections has the largest $\eta$ (suitability) largely due to its short code length.
\item LRBP has the highest level of error and wrong digits and is ranked lowest according to $\eta$. For LRBP: $E_{\textrm{total},n_p=4}<E_{\textrm{total},n_p=8}<\cdots<E_{\textrm{total},n_p=32}$.
\item LBP has the lowest level of error and wrong digits but is ranked 5 due to its long (7200 bits) code length. 
\item Assuming a total error of 1,733, the difference between $\frac{463.81}{1733}=26.76\%$ (for LBP, ranked 5) and $\frac{476.62}{1733}=27.50\%$ (for RBC$_4$, ranked 1) is rather negligible, hence amplifying the role of code length as quantified via $\eta$.
\end{itemize}
It should be noted that a comparison with numbers reported in literature for using IRMA dates cannot be made for two reasons: 1) literature around IRMA uses ``*'' as a possible digit value for ``don't know'' for which $\delta=0.5$ is used (Eq. \ref{eq:Rerror}). As no classifier is used in this paper due to different nature of the barcode-based retrieval, a modified version of the error function has been used making a comparison with classifiers impossible, and 2) a comparison with classifier is not necessary because this paper does not propose to use barcodes as an independent approach to CBIR but as a supplementary one.  

\subsection{Extra Experiment: ROI Implementation}
\begin{figure}[ht]
\begin{center}
\includegraphics[width=0.64\columnwidth]{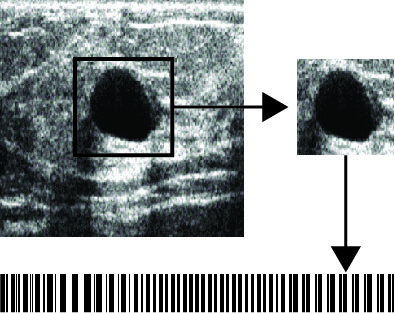} \\ 
\includegraphics[width=2cm,height=2cm]{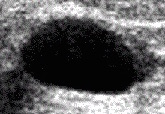}
\includegraphics[width=2cm,height=2cm]{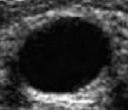} 
\includegraphics[width=2cm,height=2cm]{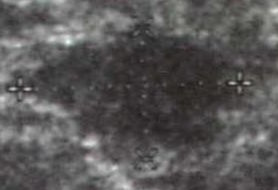}
\includegraphics[width=2cm,height=2cm]{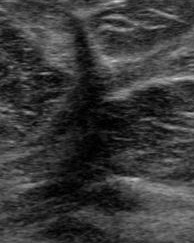}
\caption{Top: Selected ROI is annotated with a barcode for more effective intra-class retrieval. Bottom: Sample benign (left) and malignant (right) ROIs.}
\label{fig:ROIselection}
\end{center}
\end{figure}

The available datasets do not differentiate between inter- and intra-class retrieval with respect to feature detection. Assuming that we desire to use barcode annotations to encode \emph{regions of interest} (ROIs) such as tumors, barcodes could capture local characteristics of the ROI (Figure \ref{fig:ROIselection}).  To test ROI-based barcodes, 20 breast ultrasound images and their ROIs were used (Figure \ref{fig:ROIselection} bottom). Only the first hit was examined to see whether the barcodes could retrieve benign vs. malignant images correctly (hit or miss). The success rates were $\frac{7}{19}, \frac{10}{19}$ and $\frac{15}{19}$ for LRBP, LBP and RBC, respectively.

\section{Conclusions}
The idea of barcode annotations as auxiliary information beside features for medical image retrieval was proposed in this paper. Radon projections are thresholded to assemble barcodes. IRMA x-ray dataset with 14,410 images for training and testing was used to verify the performance of barcode-based image retrieval. Radon barcodes appear to provide short but expressive codes useful for medical image retrieval. Future work should investigate ROI barcodes for intra-class retrieval whereas inter-class retrieval is performed by other methods such as bag-of-feature classification paradigms. 
\newpage \newpage
\bibliographystyle{IEEEbib}
\bibliography{refs}

\end{document}